\documentclass[11pt]{article}

\usepackage[margin=1in]{geometry}
\usepackage{times}

\usepackage{amsmath,amsfonts,bm}

\def\eqref#1{equation~\ref{#1}}

\def\1{\bm{1}}

\DeclareMathAlphabet{\mathsfit}{\encodingdefault}{\sfdefault}{m}{sl}
\SetMathAlphabet{\mathsfit}{bold}{\encodingdefault}{\sfdefault}{bx}{n}

\usepackage{amsmath}
\usepackage{amssymb}
\usepackage{enumitem}
\usepackage{algorithm}
\usepackage{algpseudocode}
\usepackage{capt-of}
\usepackage{wrapfig}
\usepackage{graphicx}
\usepackage{booktabs}
\usepackage{multirow}
\usepackage{url}
\usepackage{subcaption}
\usepackage{makecell}
\usepackage[font=small]{caption}
\usepackage[dvipsnames]{xcolor}
\usepackage{natbib}
\usepackage[colorlinks=true,citecolor=purple,linkcolor=blue,urlcolor=blue]{hyperref}

\title{\textbf{Downstream-Aware Context Selection for Online In-Context Reinforcement Learning}}

\author{
Ruihan A. Li$^{1}$ \qquad
Shangtong Zhang$^{2}$ \qquad
Rohan Chandra$^{2}$\\[0.6em]
\small $^{1}$University of Illinois Urbana-Champaign\\
\small $^{2}$University of Virginia
}
\date{}

\begin{document}

\maketitle

\begin{abstract}
In-context reinforcement learning (ICRL) enables large language model agents to adapt to new environments using their interaction history without updating model parameters. However, repeatedly conditioning on growing histories can lead to substantial token cost. We propose a bounded-history context-management framework that predicts the task-dependent downstream effect of removing historical interactions to guide history selection and determine a decision-dependent context budget. Formally, our framework uses the full rolling history as a reference. The predictor evaluates removal effects, defines a deletion ordering, and applies a shared selection criterion to determine how much history to retain at each decision. We evaluate the method in closed-loop SUMO driving under held-out in-distribution, unseen-domain, and unseen-route settings, and in ScienceWorld under a continual ICRL protocol. Relative to a baseline using the full context, our method reduces total token usage by 25.7\%, 25.8\%, and 23.2\% across the three driving settings while maintaining comparable closed-loop driving performance. In ScienceWorld, it reduces total token usage by 52.1\% compared to full context and uses 30.2\% and 37.8\% fewer tokens than the Recent and Similarity baselines, respectively, while maintaining performance.
\end{abstract}

\section{Introduction}
Large language models (LLMs) have increasingly been used for sequential decision-making tasks, where previous interactions can be retained in context to support adaptation during deployment. In-context learning is useful for these tasks since it allows models to use accumulated task-specific experience and feedback during interaction to adapt behaviors. In-context reinforcement learning (ICRL) extends this idea by conditioning a policy on past state-action-reward interactions, allowing the policy to adapt during inference without updating its model parameters For instance, ICRL has enabled policy adaptation and self-improvement by providing reward feedback ~\cite{song2026reward,monea2025llmsincontextbanditreinforcement}. This capability is particularly relevant to closed-loop autonomous systems, where recent ICL/ICRL methods have been used for inference-time adaptation and decision making under changing conditions ~\cite{khurram2025promptdrivendomainadaptationendtoend, yuan2024rag}. However, as an agent continues to interact with the environment, its history grows and more information must be repeatedly included in the policy prompt.

Large context can increase computational cost ~\cite{huang2024context} and may affect adaptation in complex environments ~\cite{schmied2025retrievalaugmenteddecisiontransformerexternal}. Therefore, context selection and compression is meaningful. Existing approaches reduce long contexts through temporal abstraction, external-memory retrieval, and demonstration selection based on signals such as similarity, diversity, and learned relevance \cite{huang2024context,schmied2025retrievalaugmenteddecisiontransformerexternal,liu2022makes,levy2023diverse,scarlatos2024reticlsequentialretrievalincontext}. Recent methods further move beyond fixed-size retrieval. Context-Picker learns variable-size sufficient evidence sets using removal-based supervision \cite{zhu2026contextpickerdynamiccontextselection}, while PACE varies how much of each past memory is retained, according to predicted relevance to the next action \cite{wei-etal-2026-pace}. General prompt-compression methods such as LongLLMLingua instead reduce long inputs through token-level compression \cite{jiang-etal-2024-longllmlingua}. Removal-based influence has also been studied for retrieved contexts in question-answering and RAG settings, where importance is estimated from the performance change induced by removing a candidate context item \cite{deng2026influence}. However, existing context selection methods address two neighboring but distinct problems. Influence-based and minimal-sufficient context selectors learn which retrieved histories matter for a given query, including approaches that supervise selection through the downstream effect of removing candidate contexts. Separately, long-horizon agent methods manage evolving interaction histories through relevance-guided retrieval, multi-level compression, or learned removal. However, in online ICRL, the context is an evolving state–action–reward trajectory generated by the same policy being conditioned, and context reduction must be performed repeatedly as each action changes the history available to subsequent decisions.

\begin{wrapfigure}{r}{0.60\textwidth}
    \centering
    \includegraphics[width=\linewidth]{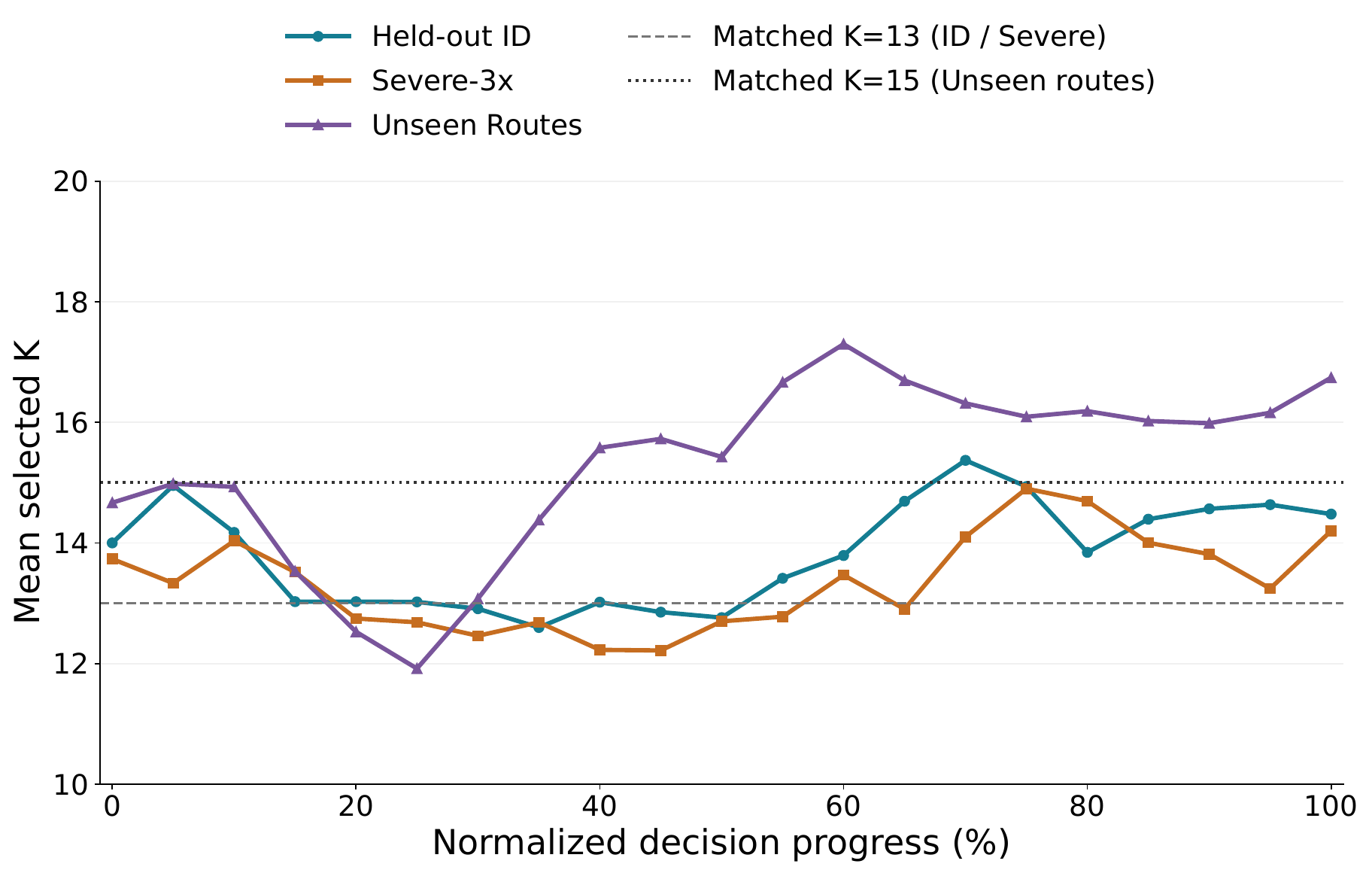}
    \caption{Mean selected context size $K$ across normalized decision
    progress under held-out ID, Severe-3x, and unseen-route evaluation.
    Dashed and dotted lines denote the matched fixed-$K$ baselines.}
    \label{fig:mean_k_by_step}
\end{wrapfigure}

We introduce a bounded-history context-selection framework for ICRL. The full rolling history serves as a reference, and a task-dependent decision-difference signal measures the downstream change induced
by replacing it with a reduced subset. Directly evaluating this quantity for many candidate subsets would require repeated downstream-policy queries, so we train a lightweight predictor to estimate the effect of context removal. The predictor then defines a deletion order, and a selection criterion then selects how much history to retain at each decision. This allows the context budget to adapt online without separately tuning a fixed history size for each condition, without requiring a fixed context cardinality as an input to the selector as demonstrated in Fig.\ref{fig:mean_k_by_step}. 

We evaluate our method in two distinct sequential decision-making domains. In a closed-loop SUMO driving environment based on the ICRL framework of ~\cite{khurram2025promptdrivendomainadaptationendtoend}. we consider held-out in-distribution episodes, an
unseen traffic domain, and unseen routes. Relative to Full-H20, our method
reduces total token usage by 23.2--25.8\% across these settings while maintaining comparable closed-loop driving performance.

We further evaluate the framework on ScienceWorld following the continual ICRL protocol of ~\cite{wang2022scienceworld,song2026reward}. Our method reduces average total token usage by 52.1\% relative to Full-H20, while achieving a return of 96.06 compared with 95.50 for
Full-H20. It also reduces token usage by 30.2\% relative to
Recent and 37.8\% relative to Similarity. A matched-budget comparison with LongLLMLingua \cite{jiang-etal-2024-longllmlingua} shows comparable ScienceWorld performance with substantially lower observed context-management overhead. These results show that the downstream-consequence-based context selection can provide effective context management across different forms of sequential
decision making and downstream supervision.

Our main contributions are:
\begin{enumerate}[leftmargin=1.5em, nosep]
    \item 
    We formulate the online context selection problem for ICRL, with the goal of reducing the token cost of a growing interaction history during inference, while maintaining downstream task performance. We formulate it around the predicted downstream consequence of removing interaction history. Unlike selecting over a fixed retrieved set, the selector operates repeatedly inside the policy–environment loop.
    
    \item 
    We propose a context management framework which predicts the effect of removing historical interactions, aggregates predictor rankings into a deletion order that also guides redundancy compression within the retained context, and adaptively selects the retained history size online based on a selection criterion. 
    
    \item We evaluate the method under two distinct environments: a closed-loop autonomous driving environment under held-out in-distribution, unseen-domain, and unseen-route settings, and ScienceWorld. Across both domains, the method reduces token usage relative to full-context and heuristic baselines while maintaining competitive downstream performance, and matches a prompt-compression baseline under a comparable token budget at substantially lower selection cost.
\end{enumerate}

\section{Related Work}
\textbf{In-Context Learning/In-Context Reinforcement Learning}
In-context learning (ICL) allows pretrained language models to adapt to new tasks using examples and demonstrations provided in the prompt without updating model parameters or fine-tuning ~\cite{brown2020language}. In reinforcement learning, Algorithm Distillation trains a causal sequence model on learning histories generated by a source RL algorithm, while the Decision-Pretrained Transformer predicts actions from a query state and an in-context dataset of past interactions \cite{laskin2022context,lee2023supervised}

More recent work studies ICRL directly with large language models. For instance, ~\cite{monea2025llmsincontextbanditreinforcement} showed that LLMs can adapt online from external reward feedback in contextual bandit problems, while ~\cite{song2026reward} introduced a multi-round ICRl prompting to guide LLM self-improvement with only numerical scalar reward. 
~\cite{chen2025filtering} studied learning-history filtering and showed that reweighting and filtering suboptimal histories during ICRL pretraining can improve downstream performance. Interaction histories and memory have also been used in LLM-based autonomous driving for shared experience and reflection, inference-time domain adaptation, risk-sensitive memory retrieval, and closed-loop reasoning and memorization \cite{jiang-etal-2024-longllmlingua,khurram2025promptdrivendomainadaptationendtoend,fu2024drive,zhou2026safedrive}

We focus on managing the accumulated context history during deployment. Rather than filtering learning histories before training or retrieving experiences using a fixed relevance criterion, we use a task-dependent estimate of the effect of context removal to construct the policy context online.

\vspace{4pt}
\textbf{Context Selection for ICL}
The performance of ICL is sensitive to the included demonstrations in the prompt. Early retrieval methods select query-relevant examples using similarity or Language Model(LM)-scored relevance \cite{liu2022makes,rubin2022learning} Some recent work also considered selection beyond direct similarity. For example, ~\cite{levy2023diverse} designed a model to select diverse demonstrations to improve compositional generalization, while ~\cite{an2023skill} proposed Skill-KNN, which selects demonstrations based on task-related skill descriptions.

More recent methods make demonstration selection more dependent on the behavior of the language model (LM). ~\cite{peng2024revisiting} indicated that demonstration quality depends on both the data and the model, and proposed a model-dependent selection method. ~\cite{wang2024demonstration} proposed Relevance-Diversity Enhanced Selection (RDES) to select demonstrations based on relevance and diversity. ~\cite{purohit2025sample} formulated Challenger Arm Sampling for Exemplar selection (CASE) framework to select from top $m$ best-arm by a linear scoring function to reduce the number of LLM evaluations needed during subset search. ~\cite{zhang-etal-2025-learning-select} proposed GENICL that leverages LLM feedback to learn demonstrations preferred by the model. Other work moves beyond independent fixed-budget selection. For instance, ~\cite{scarlatos2024reticlsequentialretrievalincontext} solved sequential example selection via Markov Decision Process (MDP) and trained example retriever using RL, while ~\cite{zhu2026contextpickerdynamiccontextselection} dynamically selects a sufficient context set through a two-stage RL framework. \cite{deng2026influence} further explore contextual influence through the performance degradation caused by removing retrieved history and train model to predict this influence at inference time. 

However, these methods improve how in-context examples are chosen,
including through sequential and adaptive selection. These methods also prove that downstream or removal-based signals can outperform semantic criteria, but build on a candidate collection associated with a query rather than repeatedly managing a trajectory generated by an online policy.

\vspace{4pt}
\textbf{Memory and Context Management}
Long-horizon language agents often use memory to retain information from previous interactions. ~\cite{huang2024context} reduce sequence length by replacing low-level action sequences with high-level decisions for interactions. MemGPT manages information across different memory levels to extend the effective context available to an LLM ~\cite{packer2023memgpt}. Other works directly reduce long prompts such as ~\cite{jiang-etal-2023-llmlingua} and ~\cite{jiang-etal-2024-longllmlingua} compress input prompts while preserving important information, reducing the token cost of long-context inference, and addressing position bias. Additionally, ~\cite{hu-etal-2025-hiagent} used subgoals formulated by LLM before each step to increase success while reducing context. MEM1 ~\cite{zhou2026mem1} used reinforcement learning to maintain a compact shared internal state that jointly supports memory consolidation and reasoning. ACON ~\cite{kang2025acon} learns compression guidelines from failure analysis in smaller models, while ~\cite{zhang2026memory} treats context deletion and insertion as learnable policy actions to optimize both information retention and task performance. More recently, PACE adjusts how much historical memory is retained according to predicted relevance for the next action \cite{wei-etal-2026-pace}, while Agent-Omit trains agents to selectively remove redundant thoughts and observations during multi-turn interactions \cite{ning2026agent}.

These methods address evolving histories, but manage them through relevance, alternative memory representations, learned agent-side deletion, or budget constraints. Different from methods that compress interaction history into a summary or a learned compact memory state, our method preserves the original state–action–reward interaction structure. It first uses a lightweight external predictor of task-dependent removal effects to repeatedly select from the evolving history, without retraining the downstream policy to perform memory selection.

\section{Methodology}

\begin{wrapfigure}[25]{R}{0.49\textwidth}
\vspace{-10pt}
\begin{minipage}{\linewidth}
\captionof{algorithm}{Adaptive Context Selection}
\vspace{-5pt}
\label{alg:adaptive_context}
\footnotesize
\hrule height 1.2pt
\vspace{2pt}
\begin{algorithmic}[1]
\Require Current query $q_t$; rolling history
$H_t=\{h_1,\ldots,h_N\}$; predictor ensemble
$\{g_m\}_{m=1}^{M}$; minimum context size $K_{\min}$;
window size $W$
\vspace{2pt}
\hrule
\vspace{2pt}
\If{$N < W$}
    \State \Return the policy prompt constructed from $q_t$ and all histories in $H_t$
\EndIf

\For{$h_i \in H_t$}
    \For{$m=1,\ldots,M$}
        \State
        $\hat y_m^{(i)}
        \gets
        g_m(q_t,H_t,H_t\setminus\{h_i\})$
    \EndFor
\EndFor

\For{$m=1,\ldots,M$}
    \State Rank deletion candidates to obtain
    $r_i^{(m)}$
\EndFor

\State $B_i \gets \sum_{m=1}^{M} r_i^{(m)}$
\State Sort $B_i$ to obtain
$o_t=(o_1,\ldots,o_N)$

\For{$K=N-1,\ldots,K_{\min}$}
    \State
    $S_K \gets
    H_t\setminus
    \{h_{o_1},\ldots,h_{o_{N-K}}\}$

    \For{$m=1,\ldots,M$}
        \State
        $\hat y_m(S_K)
        \gets
        g_m(q_t,H_t,S_K)$
    \EndFor
\EndFor

\State
$K_t^\star
\gets
\mathcal{C}\!\left(
\{\hat y_m(S_K)\}_{m,K}
\right)$ ($\mathcal{C}$ is selection rule)

\State
$S_t^\star
\gets
S_{K_t^\star}$

\State
$P_t
\gets
\mathrm{\Gamma}(q_t,\widetilde S_t^\star,\rho_t)$

\State Execute policy using $P_t$

\State Append the new interaction to $H_t$;
remove the oldest item once the window is full

\end{algorithmic}

\hrule height 1.2pt
\vspace{2pt}
\end{minipage}
\vspace{-4pt}
\end{wrapfigure}

\textbf{Problem Formulation}
At each decision time $t$, the policy receives a current query $q_t$ together with a current history rolling window $ H_t = \{h_1,\ldots,h_N\}$, where $N=|H_t|$ denotes the number of available history items. We use $K\leq N$ to denote the number of history items retained after context selection.

We treat the policy conditioned on the full history, $\pi_{\mathrm{Full}}(\cdot \mid q_t,H_t)$ as the reference policy. We maintain a fixed-capacity rolling window $H_t$ of the most recent interactions; once full, each new interaction enters the window and the oldest interaction is removed.

The goal is to reduce the total tokens used by the downstream policy while preserving its decision behavior and task performance. For a retained history subset $S \subseteq H_t$, let $\pi_S(\cdot \mid q_t, S)$ denote the policy conditioned on the current query and the retained history $S$. We measure the effect of context reduction through a task-specific
decision-difference signal
\begin{equation}
    D_t(S) = \mathcal{D}(q_t,H_t,S),
    \label{eq:decision_gap}
\end{equation}
where smaller values indicate that the selected context $S$ better
preserves the behavior induced by the full history. The specific form of
$\mathcal{D}$ depends on the downstream task: it can measure either
distributional changes in the policy output or degradation in the
resulting task outcome.

Therefore, the selected context is not necessarily the most recent or the most state-similar history, but a compact set with a small decision gap relative to the reference policy. However, directly evaluating the decision gap for many subsets would require repeated LLM queries at every decision step, and the number of possible subsets grows exponentially with $N$. We therefore learn a predictor that produces a behavior-aware deletion ordering over the full history. Let
\[
o_t=
\left(
o_{t,1},\ldots,o_{t,N}
\right)
\]
denote the order in which history items are removed. For each history item $h_i$, let $\rho_{t,i}$ denote its position in this deletion order, \[
\rho_{t,i}
=
\operatorname{pos}_{o_t}(i), \quad \rho_t = (\rho_{t,1}, \ldots, \rho_{t,N}),
\]
where a larger $\rho_{t,i}$ means that the history item is removed later. 

The learned deletion order is used in two places. First, it determines which historical items are kept and therefore the adaptive context size.
Second, after selecting the context, the same order is reused
to reduce repeated contextual information within those histories. For
each repeated value, one retained occurrence keeps the complete text,
while the other occurrences are replaced by references to that
copy. Thus, the final prompt depends jointly on which interactions are retained and how repeated information is represented within them.

\subsection{Decision Gap Predictor}
\paragraph{Decision gap supervision}

The predictor is trained to estimate the effect of removing selected context subsets.

For autonomous driving, let $\mathcal{A}$ denote the discrete action space. For a prompt $p$, let $\ell_a(p)$ denote the log probability of action token $a$ at the first action position. We construct a normalized action distribution over the valid action set as \[ \pi_p(a) = \frac{\exp(\ell_a(p))} {\sum_{b\in\mathcal{A}}\exp(\ell_b(p))}\] This distribution represents the policy's decision under prompt $p$ and is used to compute the decision gap between the full-context and selected-context prompts. For a current query $q_t$ and selected context $S$, the predictor target is 
\begin{equation}
    y_t(S) = \log\left( 1+ D_{\mathrm{KL}} \left( \pi_{\mathrm{Full}}(\cdot\mid q_t,H_t) \,\Vert\, \pi_{S}(\cdot\mid q_t,S) \right) \right).
    \label{eq:driving_target}
\end{equation}

Thus, the supervision measures the effect of context removal on the downstream action distribution rather than semantic similarity between the current state and a historical experience.

For ScienceWorld, we instead measure the terminal outcome
degradation induced by context reduction:
\begin{equation}
    y_t^{\mathrm{SW}}(S)
=
\frac{
\bar R_{\mathrm{Full}}
-
\bar R_S
}{200},
\label{eq:sw_target}
\end{equation}
where $\bar R_{\mathrm{Full}}$ and $\bar R_S$ denote the mean terminal
scores obtained using the full and selected contexts, respectively.

Thus, in both domains, the supervision directly measures the effect of
context removal on downstream behavior or task performance.

\vspace{2pt}
\textbf{Representation}
We define the current query as $q_t$ and each history item as $h_i$. They are encoded by a frozen text embedding model, $e_t = E(q_t), e_i = E(h_i)$. Each history item is additionally associated with a structured task-dependent feature vector $r_i$ containing decision-time information. The semantic and structured features are contextualized to obtain representations $u_t$ for the current query and $u_i$ for the history items.
The Transformer inputs are constructed as
\[
x_t=W_q e_t+p_0,
\]
\[
x_i
=
W_c\!\left([\bar{e}_i;\tilde{r}_i]\right)
+
E_{\mathrm{status}}(s_i)
+
p_i
\]
where $W_q$ and $W_c$ are learned projections, $E_{\mathrm{status}}$ is a learned status embedding, and $p_i$ denotes the learned embedding of the corresponding sequence position. The construction of $\tilde{e}_i$ and $\tilde{r}_i$ is task-specific. In autonomous driving, histories removed in earlier deletion steps are masked, while the current deletion candidate remains visible. In ScienceWorld, the original attempt representations remain available and the status embedding indicates whether each attempt is selected or excluded.

The resulting sequence is processed by a Transformer encoder with one layer, four attention heads, and hidden dimension 64:
\[
(u_t,u_1,\ldots,u_{N})
=
\operatorname{Transformer}(x_t,x_1,\ldots,x_{N}),
\]
where $u_t$ and $u_i$ are the contextualized representations of the current query and history item $h_i$, respectively.

For a selected context $S$, we construct a candidate
representation $\Phi_t(H_t,S)$ from the contextualized history
representations. It summarizes the relationship between the full
history and the selected context, together with task-dependent decision information when available. The predictor then
estimates the task-specific decision difference as
\begin{equation}
    \hat y_t(S)
=
g_\theta\!\left(u_t,\Phi_t(H_t,S)\right).
\label{eq:predictor}
\end{equation}

Task-specific representation are provided in Appendix \ref{app:representation}

\vspace{2pt}
\textbf{Training objective}
We train the predictor with both regression and ranking terms.
Firstly, for a candidate context \(S\), the prediction target is
\[
y_t(S) = \log\left(1 + D_t(S)\right).
\] for autonomous driving and 
\[
y_t^{(S)}
=
\frac{
\bar R_{\mathrm{Full}}
-
\bar R_S
}{200},
\] for ScienceWorld. 

The regression loss is
\[
\mathcal{L}_{\mathrm{reg}}
=
\operatorname{Huber}\left(\hat{y}_t(S), y_t(S)\right).
\]
Secondly, for two selected candidates \(i\) and \(j\) from the same parent context (or between equal-cardinality candidates from the same history pool in ScienceWorld.), we also match their target difference,
\[
\mathcal{L}_{\Delta}
=
\operatorname{Huber}(\hat{y}_i-\hat{y}_j,\; y_i-y_j),
\]
Finally, we add a pairwise ranking loss that prefers the candidate with the smaller true decision gap,
\[
\mathcal{L}_{\mathrm{rank}}
=
|y_i-y_j|\,
\operatorname{BCEWithLogits}
\left(
\hat{y}_j-\hat{y}_i,\;
\mathbf{1}[y_i<y_j]
\right).
\]

The final objective is
\begin{equation}
    \mathcal{L}
=
\mathcal{L}_{\mathrm{reg}}
+
0.5\mathcal{L}_{\Delta}
+
0.25\mathcal{L}_{\mathrm{rank}}.
\label{eq:training_objective}
\end{equation}

This gives the predictor an absolute decision gap target while also training the ordering that will be used by the selector. 

\textbf{Predictor training.}
We train a five-member task-specific predictor ensemble in each domain.
For driving, supervision is the log-transformed KL divergence between the
full- and reduced-context action distributions, with labels generated using query and history
representations use frozen 3,072-dimensional \texttt{text-embedding-3-large} embeddings.
The driving predictor is trained on 1,956 labeled masks,
with a disjoint 89-mask split used for checkpoint selection.
For ScienceWorld, supervision is the normalized terminal-score degradation
between full and reduced contexts, estimated using three repeated rollouts
per condition under the frozen rollout protocol. The ScienceWorld predictor
uses 1,540 training labels and label-disjoint 150 checkpoint-selection labels, with frozen \texttt{all-MiniLM-L6-v2} embeddings.
Both predictors are optimized with AdamW using the regression, difference,
and ranking losses in Eq.~(5); complete training parameters are provided in Appendix \ref{app:predictor}.

\subsection{Deletion-Structured Context Construction}
\paragraph{Deletion Ranking}
The deletion candidates are ranked based on predicted decision gap from smallest to largest by each predictor. Let \(r_i^{(m)}\) be the rank of candidate \(i\) under member \(m\). We combine the five rankings with a Borda score,
\begin{equation}
    B_i = \sum_{m=1}^{5} r_i^{(m)}.
\label{eq:borda}
\end{equation}

Candidates with smaller \(B_i\) are deleted earlier. If two candidates have the same score, we break the tie using the mean predicted gap, then the history length, and finally the history index. This gives one fixed deletion order $o_t = (o_1,o_2,\ldots,o_{N})$

The items near the end of this order will be deleted last.

\textbf{Adaptive History Selection} Based on the deletion order \(o_t\), we construct a sequence for the contexts. For a context size \(K\), we remove the first \(N-K\) items in the deletion order:
\begin{equation}
    S_K
=
H_t \setminus
\{h_{o_1},\ldots,h_{o_{N-K}}\},
\qquad
K \in \{10,\ldots,19\}.
\label{eq:nested_context}
\end{equation}

Though the deletion ranking is fixed for the current decision, the predicted gap will be recomputed at each context size while excluding the items that have already been removed. For example, when evaluating \(S_{18}\), the predictor knows that the first-ranked item has already been removed and evaluates the next deletion under this reduced context

For each candidate context $S_K$, the five predictor outputs are
evaluated independently. We take the average of the five predictors: \[
\bar{y}_t(S_K)
=
\frac{1}{5}
\sum_{m=1}^{5}
\hat{y}_t^{(m)}(S_K).
\] 

A candidate context passes the predicted-gap threshold if
\begin{equation}
\begin{cases}
\displaystyle
\bar{y}_t(S_K) \leq \log(1+\tau),
& \text{Driving}, \\[4pt]
\displaystyle
\frac{1}{5}\sum_{m=1}^{5}
\left[
\hat{y}_t^{(m)}(S_K)
-
\min_{j\in\{10,\ldots,20\}}
\hat{y}_t^{(m)}(S_j)
\right]
\leq \tau,
& \text{ScienceWorld},
\end{cases}
\label{eq:feasibility}
\end{equation}
where $\tau=0.05$.

We set $K_{\min}=10$ as a fixed search hyperparameter and then evaluate candidate context sizes from \(K=19\) to \(K=10\) and select the smallest context that satisfies the threshold mentioned above. 

If none of the reduced contexts satisfies the threshold, it will automatically keep full window ($N=20$). Then, the same deletion order remains part of the context representation and determines how repeated contextual values are allocated within the retained chunks. Since we do not assume that the predicted gap changes monotonically with $K$, an intermediate context size that fails the threshold does not stop the search. 

\textbf{Window-Size}
We choose $W=20$ as the capacity of the rolling history window. The window is updated throughout an episode: before it is full, all available interactions are retained; once $N=W$, each new interaction enters the window while the oldest one is removed, and adaptive selection is applied over the current rolling history. Each driving episode contain approximately 37--42 decisions on average across the evaluated settings, while ScienceWorld contains 40 interaction rounds. A substantially larger $W$ would leave relatively few decisions after the history window becomes full. With $W=20$, roughly half of trajectories remains available for selection-active evaluation.

\textbf{Order-guided Redundancy Compression} After selecting the context chunks, repeated contextual information
may still appear across multiple retained interactions. We
therefore apply an additional compression step using the
same deletion order. This step reduces repeated text within
the selected history by storing an identical contextual value
once. The repeated text itself is omitted at that occurrence,
while the reference allows the original value to be recovered
from its retained source. Here, a reference is a textual marker inserted in place of a repeated value that identifies where the complete value is
retained. Detailed field definitions and compression rules are provided in Appendix \ref{app:lorc}

\section{Results and Analysis}
\textbf{Environment Setup} We test our method under two environments. Firstly, we evaluate in a closed-loop SUMO driving environment built on the framework introduced in ~\cite{khurram2025promptdrivendomainadaptationendtoend}, which is based on the LimSim++ closed-loop driving platform ~\cite{fu2024limsim++}. We define the discrete driving action space $\mathcal{A}=\{1,2,3,4,8\}$, corresponding to acceleration, deceleration, left lane change, right lane change, and idle, respectively. All methods are tested on a rolling window of the \(W=20\) most recent interactions. Before the window is full, all available histories are kept; once full, the window advances as new interactions are recorded and context selection is applied at each subsequent decision. Following ~\cite{khurram2025promptdrivendomainadaptationendtoend}, predictor development uses nine seen domains defined by three weather conditions and three traffic densities. We evaluate on (i) 23 held-out in-distribution episodes, (ii) 13 episodes from the unseen Severe-3x domain, and (iii) 27 unseen-route conditions covering three held-out routes, three seen traffic domains, and three random seeds. 

Secondly, we evaluate our method under ScienceWorld, following the continual ICRL evaluation protocol of ~\cite{song2026reward}. It is a text-based interactive environment whose action space includes high-level operational commands for movement between locations, object manipulation, and scientific experimentation. We include it to test whether the proposed context-selection mechanism transfers beyond the closed-loop driving setting, using a different form of downstream supervision. We use two initial attempts followed by 40 interaction rounds, with exploration and exploitation instructions alternating across rounds. We evaluate randomly selected 18 of the 30 ScienceWorld tasks and a maximum of 15 environment steps per episode. Since our framework targets bounded-history context selection, we maintain a candidate window of at most the 20 most recent experiences; our method adaptively selects and compresses this history, while the Full baseline retains all experiences within the H20 window. Recent and Similarity use approximately matched context cardinality as ours. The downstream policy model is GPT-4o for driving and GPT-4.1 mini for ScienceWorld.

\vspace{4pt}
\textbf{Baselines}
We also compare our method with three context baselines: Full H20, Similarity-K, and Recent-K. Compression-controlled baselines: Recent+Redundancy Compression (Recent + LORC). We additionally compare our method with LongLLMLingua, a general long-context prompt-compression baseline, on ScienceWorld \cite{jiang-etal-2024-longllmlingua}. Detailed definitions are shown in ~\ref{app:baseline}

\vspace{4pt}
\textbf{Evaluation Metrics}
We evaluate baselines performance under several metrics: 
\begin{enumerate}[leftmargin=1.5em, nosep]
    \item We evaluate driving performance using \textbf{Safety, Comfort, and Efficiency} ~\cite{khurram2025promptdrivendomainadaptationendtoend,fu2024limsim++}. All three driving metrics range from 0 to 1, with higher values indicating better performance. Safety measures collision-related risk, Comfort measures trajectory smoothness based on acceleration and jerk, and Efficiency measures driving efficiency with respect to traffic conditions and road-speed constraints. Detailed metric definitions and aggregation follow the original evaluation protocol in ~\cite{khurram2025promptdrivendomainadaptationendtoend,fu2024limsim++}

    \item For ScienceWorld, we compare our method with baselines via \textbf{Return}. We follow the same evaluation metrics as \cite{song2026reward}. Score achieved has been averaged across the evaluated tasks for all-round(Round 1-40)/post-selection(Round 19-40). The two bootstrap trials are excluded.

    \item We also evaluate our method under tokens. \textbf{Post-selection tokens} measure the total number of tokens provided to the downstream policy over decision steps and generated output tokens after the rolling history reaches its full capacity (\(N=W=20\)) and context selection becomes active. The metric includes the current query and the selected/compressed history, and is accumulated over the post-H20 portion of each episode before averaging across episodes.

\textbf{Total Tokens} measure the total number of tokens consumed by the downstream policy over the entire episode, including both input and generated output tokens across all decision steps:
\[
T_{\mathrm{total}} = T_{\mathrm{in}} + T_{\mathrm{out}}.
\]
\end{enumerate}

\begin{table*}[!t]
\centering
\caption{
Evaluation results across the held-out in-distribution, unseen OOD, and unseen-route settings.
Token counts are rounded to the nearest thousand (K), and performance scores to two decimals.
Lower token usage is better, while higher Safety, Comfort, and Efficiency scores are better.
Bold indicates the best value based on the unrounded results.
}
\label{tab:driving_results}
\vspace{-4pt}

\begingroup
\scriptsize

\setlength{\tabcolsep}{4pt}

\renewcommand{\arraystretch}{0.78}

\setlength{\aboverulesep}{1pt}
\setlength{\belowrulesep}{1pt}
\setlength{\abovetopsep}{0pt}
\setlength{\belowbottomsep}{0pt}

\begin{tabular*}{0.96\textwidth}{
@{\extracolsep{\fill}}
llcccccc
@{}}
\toprule
\makecell{\textbf{Evaluation}\\\textbf{Type}}
& \textbf{Method}
& \textbf{Mean $K$}
& \makecell{\textbf{Post-selection}\\\textbf{Tokens$\downarrow$}}
& \textbf{Avg Total$\downarrow$}
& \textbf{Safety$\uparrow$}
& \textbf{Comfort$\uparrow$}
& \textbf{Efficiency$\uparrow$} \\
\midrule

\multirow{4}{*}{\makecell{Held-out\\ID}}
& Full-H20
& $20.00$
& $130$K
& $218$K
& $0.88$
& $0.76$
& $0.79$ \\

& Recent-13
& $13.00$
& $92$K
& $181$K
& $\mathbf{0.88}$
& $0.76$
& $0.80$ \\

& Similarity-13
& $13.00$
& $91$K
& $180$K
& $0.88$
& $\mathbf{0.77}$
& $0.80$ \\

& \textbf{Ours}
& $13.63$
& $\mathbf{73}$K
& $\mathbf{162}$K
& $0.88$
& $0.76$
& $\mathbf{0.81}$ \\

\midrule

\multirow{4}{*}{\makecell{Severe-\\3x}}
& Full-H20
& $20.00$
& $151$K
& $247$K
& $0.82$
& $0.78$
& $0.83$ \\

& Recent-13
& $13.00$
& $103$K
& $201$K
& $0.87$
& $0.77$
& $0.84$ \\

& Similarity-13
& $13.00$
& $103$K
& $200$K
& $\mathbf{0.87}$
& $\mathbf{0.78}$
& $\mathbf{0.84}$ \\

& \textbf{Ours}
& $13.02$
& $\mathbf{86}$K
& $\mathbf{184}$K
& $0.86$
& $0.78$
& $0.83$ \\

\midrule

\multirow{4}{*}{\makecell{Unseen\\Routes}}
& Full-H20
& $20.00$
& $122$K
& $201$K
& $\mathbf{0.98}$
& $0.74$
& $0.87$ \\

& Recent-15
& $15.00$
& $96$K
& $175$K
& $0.98$
& $\mathbf{0.75}$
& $\mathbf{0.88}$ \\

& Similarity-15
& $15.00$
& $98$K
& $177$K
& $0.98$
& $0.74$
& $0.88$ \\

& \textbf{Ours}
& $15.05$
& $\mathbf{75}$K
& $\mathbf{154}$K
& $0.98$
& $0.74$
& $0.87$ \\

\bottomrule
\end{tabular*}

\endgroup
\vspace{-6pt}
\end{table*}

\subsection{Autonomous-Driving}
\textbf{Held-out In-distribution} 
For in-distribution test, our method substantially reduces context cost while maintaining driving performance. Compared with Full-H20, Ours reduces total token usage by 25.7$\%$, with a paired-bootstrap 95\% confidence interval of \textbf{[$-29.2\%$, $-22.2\%$]}, using an average of 13.63 histories. For Recent-13 and Similarity-13, which use comparable history budgets, Ours reduces total tokens by 10.3$\%$ and 9.7$\%$, respectively. These reductions are also consistent under paired bootstrap, with 95\% confidence intervals of \textbf{[$-13.8\%$, $-6.6\%$]} and \textbf{[$-13.8\%$, $-5.4\%$]}. 

\textbf{Unseen Domain OOD} Under out-of-distribution testing, our method continues to reduce token usage relative to the baselines. Compared with Full-H20, Ours reduces average total token usage by 25.8\%, from $247{,}490$ to $183{,}521$, with a route-clustered 95\% confidence interval of $[-31.2\%, -19.4\%]$, while using an average context size of $K = 13.02$. Ours also uses fewer total tokens than Recent-13 and Similarity-13 by 8.6\% and 8.5\%, respectively. Driving performance remains comparable overall: relative to Full-H20, Ours improves Safety from $0.8211$ to $0.8625$, maintains the same Comfort score of $0.7759$, and yields a slightly lower Efficiency score ($0.8256$ vs.\ $0.8268$). These results suggest that the complete context-selection mechanism remains effective under the unseen traffic domain.

\textbf{Unseen Routes OOD}
We further evaluate the method on three routes that are not used in predictor training, covering three traffic domains (Clear 1$\times$, slightly inclement 2$\times$, and moderately inclement 3$\times$), and three random seeds for a total of 27 matched conditions. These routes follow distinct edge sequences and cover complementary maneuver patterns: straight driving, a required left lane change followed by a left turn, and a required right lane change. Compared with Full-H20, Ours reduces the average post-selection tokens from $121{,}753$ to $75{,}138$, a 38.3\% reduction, while total token usage decreases by 23.2\%. Ours uses fewer total tokens in all 27/27 matched route-domain-seed conditions, with a paired-bootstrap 95\% confidence interval of [$-24.98\%$, $-21.40\%$]. Although Ours retains slightly more history on average ($\bar{K}=15.05$) than Recent-15 and Similarity-15, it uses 21.6\% and 23.0\% fewer post-selection tokens, and 11.9\% and 12.7\% fewer total tokens, respectively, with paired-bootstrap 95\% confidence intervals of [$-13.59\%$, $-10.12\%$] and [$-14.71\%$, $-10.67\%$]. Driving performance also remains close to Full-H20, with only small changes in Safety, Comfort, and Efficiency. Paired-bootstrap confidence intervals for performance are reported in Appendix \ref{app:ci}.

\begin{wraptable}[10]{r}{0.68\textwidth}
\vspace{-12pt}
\centering
\caption{
ScienceWorld evaluation. Late-round statistics are averaged over rounds 19--40, when context selection is active for our method, whereas all-round statistics are averaged over rounds 1--40.
Because histories grow over the episodes, late-round token usage can exceed the all-round average. ``Total'' includes both input and output tokens. "Return" are average over all-round.
}
\label{tab:main_results}
\vspace{-4pt}

\scriptsize
\renewcommand{\arraystretch}{0.92}
\setlength{\tabcolsep}{3pt}

\begin{tabular*}{\linewidth}{@{\extracolsep{\fill}}lcccc@{}}
\toprule
\textbf{Method}
& \textbf{Mean $K$}
& \makecell{\textbf{Late-Round}\\\textbf{Total/Trial$\downarrow$}}
& \makecell{\textbf{Avg Total}\\\textbf{/Trial$\downarrow$}}
& \textbf{Return$\uparrow$} \\
\midrule

Full-H20
& $20.00$
& $352{,}809$
& $282{,}879$
& $95.50$ \\

Recent-12
& $12.00$
& $202{,}710$
& $194{,}188$
& $\mathbf{96.06}$ \\

Similarity-12
& $12.00$
& $227{,}433$
& $217{,}706$
& $94.78$ \\

\textbf{Ours}
& $12.70$
& $\mathbf{87{,}100}$
& $\mathbf{135{,}470}$
& $\mathbf{96.06}$ \\

\bottomrule
\end{tabular*}

\vspace{-8pt}
\end{wraptable}

\subsection{ScienceWorld Evaluation}
As shown in Table~\ref{tab:main_results}, Ours achieves the lowest token usage while matching the observed running-max return of the recency-based baseline. Ours and Recent both achieve a return of 96.06, while Ours reduces average total token usage by 30.2\%. Relative to Full, Ours reduces total tokens by 52.1\% and achieves a slightly higher observed return (96.06 vs.\ 95.50). Compared with Similarity, it reduces total tokens by 37.8\% while obtaining a higher observed return (96.06 vs.\ 94.78). Task-level paired bootstrap analysis yields mean total-token differences per formal trial of $-147.4$k relative to Full (95\% CI [$-215.5$k, $-95.1$k]), $-58.7$k relative to Recent ([$-80.8$k, $-37.2$k]), and $-82.2$k relative to Similarity ([$-127.8$k, $-44.7$k]). These results indicate substantial context-cost reductions while maintaining competitive observed task performance. We additionally report mean return, post-selection return (round 19-40), trial-level return trajectories and running-max performance in Appendix \ref{app:sw_performance}. During the selection-active phase, Ours maintains competitive task performance, achieving a post-selection running-max/mean return of 95.94/74.22 compared with 95.50/74.34 for Full reference policy, 95.78/76.02 for Recent, 94.78/69.37 for Similarity.

\subsection{Controlled Selection and Budgeting Baselines}
We additionally apply the same learned order-
guided redundancy compression (LORC) to the Recent-K baseline. It retains the \(K\) most recent interactions while using our learned deletion order during compression. We report post-selection total
tokens to isolate the selection-active phase. A redundancy-controlled Recent-$K$ comparison shows larger gains under distribution shift: Ours reduces post-selection token usage by 12.7\% on Severe-3x and 7.8\% on unseen routes. A similar controlled advantage is observed in ScienceWorld, where Ours reduces post-selection by 7.6\%. 

We further construct matched-$K$ fixed-budget baseline using the same method as ours, where the fixed context size is chosen to match the mean retained history size observed for Ours in each setting instead of selecting independently. At comparable average context cardinalities, the variants yield similar operating points on held-out ID and unseen routes, while Ours uses 7.38\% fewer post-selection tokens on Severe-3x. Detailed results refer to Appendix \ref{app:redundancy}

\subsection{Comparison with General Prompt Compression}
\begin{wraptable}{r}{0.49\textwidth}
\vspace{-10pt}
\centering
\caption{ScienceWorld comparison with LongLLMLingua.}
\label{tab:longllm_results}
\vspace{-5pt}

\scriptsize
\renewcommand{\arraystretch}{0.95}

\begin{tabular*}{\linewidth}{@{\extracolsep{\fill}}rcc@{}}
\toprule
\textbf{Metric}
& \textbf{Ours}
& \textbf{LongLLMLingua} \\
\midrule

Avg Total / Trial $\downarrow$
& $135{,}470$
& $\mathbf{131{,}467}$ \\

Return $\uparrow$
& $\mathbf{96.060}$
& $94.780$ \\

History Proc. Latency (s) $\downarrow$
& $\mathbf{1.244}$
& $27.715$ \\

\bottomrule
\end{tabular*}

\vspace{-8pt}
\end{wraptable}
We additionally compare with LongLLMLingua under an approximately matched historical-context budget. LongLLMLingua achieves all-round/post-selection return of 94.78/94.67, compared with 96.06/95.94 for Ours. However, its official compression core requires 27.7 s per mature decision on average in our ScienceWorld setting, whereas the complete context-management pipeline of Ours requires 1.24 s on a two-thread CPU. Detailed calibration, task-level results, and latency protocols are provided in Appendix \ref{app:longllm}.

\subsection{Feasibility Threshold Sensitivity Analysis}
We additionally perform an offline sensitivity analysis over post selection driving decisions by recomputing context selection and input-token usage on the evaluation trajectories. Across all three settings, increasing \(\tau\) consistently reduces context budget and post-selection input tokens, while the threshold \(\tau=0.05\) remains away from either budget extreme, supporting that the feasibility threshold provides controllable adaptive budgeting. Detailed results refer to Appendix \ref{app:sensitivity} 

\section{Conclusion}
In this work, we study online context selection for in-context reinforcement learning and propose a method for reducing accumulated interaction context during deployment. The method learns the effect of context reduction, constructs a deletion ordering, adaptively selects
the retained history size, and reuses the same ordering to reduce
redundant information within the retained context.

In closed-loop SUMO driving, our method reduces total token usage by
23.2--25.8\% relative to the full-context policy across held-out
in-distribution, unseen-domain, and unseen-route evaluations while
maintaining comparable driving performance. In ScienceWorld, it reduces
total token usage by 52.1\% relative to Full and by 30.2\% and 37.8\%
relative to Recent and Similarity, respectively. Overall, across both domains, the learned selector reduces context cost while maintaining comparable downstream performance.

However, our marginal deletion ranking does not explicitly model higher-order dependencies among multiple interaction histories. Consequently, the nested subsets defined by the deletion order may not contain the globally best subset at every context size. Extending the selector to capture such interactions is a direction for future work.

\bibliographystyle{plainnat}
\bibliography{reference}

@misc{schmied2025retrievalaugmenteddecisiontransformerexternal,
      title={Retrieval-Augmented Decision Transformer: External Memory for In-context RL}, 
      author={Thomas Schmied and Fabian Paischer and Vihang Patil and Markus Hofmarcher and Razvan Pascanu and Sepp Hochreiter},
      year={2025},
      eprint={2410.07071},
      archivePrefix={arXiv},
      primaryClass={cs.LG},
      url={https://arxiv.org/abs/2410.07071}, 
}

@misc{scarlatos2024reticlsequentialretrievalincontext,
      title={RetICL: Sequential Retrieval of In-Context Examples with Reinforcement Learning}, 
      author={Alexander Scarlatos and Andrew Lan},
      year={2024},
      eprint={2305.14502},
      archivePrefix={arXiv},
      primaryClass={cs.CL},
      url={https://arxiv.org/abs/2305.14502}, 
}

@misc{zhu2026contextpickerdynamiccontextselection,
      title={Context-Picker: Dynamic context selection using multi-stage reinforcement learning}, 
      author={Siyuan Zhu and Chengdong Xu and Kaiqiang Ke and Chao Yu},
      year={2026},
      eprint={2512.14465},
      archivePrefix={arXiv},
      primaryClass={cs.AI},
      url={https://arxiv.org/abs/2512.14465}, 
}

@misc{khurram2025promptdrivendomainadaptationendtoend,
      title={Prompt-Driven Domain Adaptation for End-to-End Autonomous Driving via In-Context RL}, 
      author={Aleesha Khurram and Amir Moeini and Shangtong Zhang and Rohan Chandra},
      year={2025},
      eprint={2511.12755},
      archivePrefix={arXiv},
      primaryClass={cs.RO},
      url={https://arxiv.org/abs/2511.12755}, 
}

@inproceedings{fu2024limsim++,
  title={Limsim++: A closed-loop platform for deploying multimodal llms in autonomous driving},
  author={Fu, Daocheng and Lei, Wenjie and Wen, Licheng and Cai, Pinlong and Mao, Song and Dou, Min and Shi, Botian and Qiao, Yu},
  booktitle={2024 IEEE Intelligent Vehicles Symposium (IV)},
  pages={1084--1090},
  year={2024},
  organization={IEEE}
}

@article{brown2020language,
  title={Language models are few-shot learners},
  author={Brown, Tom and Mann, Benjamin and Ryder, Nick and Subbiah, Melanie and Kaplan, Jared D and Dhariwal, Prafulla and Neelakantan, Arvind and Shyam, Pranav and Sastry, Girish and Askell, Amanda and others},
  journal={Advances in neural information processing systems},
  volume={33},
  pages={1877--1901},
  year={2020}
}

@article{laskin2022context,
  title={In-context reinforcement learning with algorithm distillation},
  author={Laskin, Michael and Wang, Luyu and Oh, Junhyuk and Parisotto, Emilio and Spencer, Stephen and Steigerwald, Richie and Strouse, DJ and Hansen, Steven and Filos, Angelos and Brooks, Ethan and others},
  journal={arXiv preprint arXiv:2210.14215},
  year={2022}
}

@article{lee2023supervised,
  title={Supervised pretraining can learn in-context reinforcement learning},
  author={Lee, Jonathan and Xie, Annie and Pacchiano, Aldo and Chandak, Yash and Finn, Chelsea and Nachum, Ofir and Brunskill, Emma},
  journal={Advances in Neural Information Processing Systems},
  volume={36},
  pages={43057--43083},
  year={2023}
}

@misc{monea2025llmsincontextbanditreinforcement,
      title={LLMs Are In-Context Bandit Reinforcement Learners}, 
      author={Giovanni Monea and Antoine Bosselut and Kianté Brantley and Yoav Artzi},
      year={2025},
      eprint={2410.05362},
      archivePrefix={arXiv},
      primaryClass={cs.CL},
      url={https://arxiv.org/abs/2410.05362}, 
}

@inproceedings{song2026reward,
  title={Reward is enough: Llms are in-context reinforcement learners},
  author={Song, Kefan and Moeini, Amir and Wang, Peng and Gong, Lei and Chandra, Rohan and Zhang, Shangtong and Qi, Yanjun},
  booktitle={International Conference on Learning Representations},
  volume={2026},
  pages={112747--112770},
  year={2026}
}

@article{chen2025filtering,
  title={Filtering learning histories enhances in-context reinforcement learning},
  author={Chen, Weiqin and Zhang, Xinjie and Subramanian, Dharmashankar and Paternain, Santiago},
  journal={arXiv preprint arXiv:2505.15143},
  year={2025}
}

@inproceedings{liu2022makes,
  title={What makes good in-context examples for GPT-3?},
  author={Liu, Jiachang and Shen, Dinghan and Zhang, Yizhe and Dolan, William B and Carin, Lawrence and Chen, Weizhu},
  booktitle={Proceedings of Deep Learning Inside Out (DeeLIO 2022): The 3rd workshop on knowledge extraction and integration for deep learning architectures},
  pages={100--114},
  year={2022}
}

@inproceedings{rubin2022learning,
  title={Learning to retrieve prompts for in-context learning},
  author={Rubin, Ohad and Herzig, Jonathan and Berant, Jonathan},
  booktitle={Proceedings of the 2022 conference of the North American chapter of the association for computational linguistics: human language technologies},
  pages={2655--2671},
  year={2022}
}

@inproceedings{levy2023diverse,
  title={Diverse demonstrations improve in-context compositional generalization},
  author={Levy, Itay and Bogin, Ben and Berant, Jonathan},
  booktitle={Proceedings of the 61st Annual Meeting of the Association for Computational Linguistics (Volume 1: Long Papers)},
  pages={1401--1422},
  year={2023}
}

@inproceedings{an2023skill,
  title={Skill-based few-shot selection for in-context learning},
  author={An, Shengnan and Zhou, Bo and Lin, Zeqi and Fu, Qiang and Chen, Bei and Zheng, Nanning and Chen, Weizhu and Lou, Jian-Guang},
  booktitle={Proceedings of the 2023 Conference on Empirical Methods in Natural Language Processing},
  pages={13472--13492},
  year={2023}
}

@inproceedings{peng2024revisiting,
  title={Revisiting demonstration selection strategies in in-context learning},
  author={Peng, Keqin and Ding, Liang and Yuan, Yancheng and Liu, Xuebo and Zhang, Min and Ouyang, Yuanxin and Tao, Dacheng},
  booktitle={Proceedings of the 62nd Annual Meeting of the Association for Computational Linguistics (Volume 1: Long Papers)},
  pages={9090--9101},
  year={2024}
}

@article{wang2024demonstration,
  title={Demonstration selection for in-context learning via reinforcement learning},
  author={Wang, Xubin and Wu, Jianfei and Yuan, Yichen and Cai, Deyu and Li, Mingzhe and Jia, Weijia},
  journal={arXiv preprint arXiv:2412.03966},
  year={2024}
}

@article{purohit2025sample,
  title={Sample efficient demonstration selection for in-context learning},
  author={Purohit, Kiran and Venktesh, V and Bhattacharya, Sourangshu and Anand, Avishek},
  journal={arXiv preprint arXiv:2506.08607},
  year={2025}
}

@inproceedings{zhang-etal-2025-learning-select,
    title = "Learning to Select In-Context Demonstration Preferred by Large Language Model",
    author = "Zhang, Zheng  and
      Lan, Shaocheng  and
      Song, Lei  and
      Bian, Jiang  and
      Li, Yexin  and
      Ren, Kan",
    editor = "Che, Wanxiang  and
      Nabende, Joyce  and
      Shutova, Ekaterina  and
      Pilehvar, Mohammad Taher",
    booktitle = "Findings of the Association for Computational Linguistics: ACL 2025",
    month = jul,
    year = "2025",
    address = "Vienna, Austria",
    publisher = "Association for Computational Linguistics",
    url = "https://aclanthology.org/2025.findings-acl.592/",
    doi = "10.18653/v1/2025.findings-acl.592",
    pages = "11345--11360",
    ISBN = "979-8-89176-256-5"
}

@article{packer2023memgpt,
  title={Memgpt: Towards llms as operating systems},
  author={Packer, Charles and Wooders, Sarah and Lin, Kevin and Fang, Vivian and Patil, Shishir G and Stoica, Ion and Gonzalez, Joseph E},
  journal={arXiv preprint arXiv:2310.08560},
  year={2023}
}

@inproceedings{jiang-etal-2023-llmlingua,
    title = "{LLML}ingua: Compressing Prompts for Accelerated Inference of Large Language Models",
    author = "Jiang, Huiqiang  and
      Wu, Qianhui  and
      Lin, Chin-Yew  and
      Yang, Yuqing  and
      Qiu, Lili",
    editor = "Bouamor, Houda  and
      Pino, Juan  and
      Bali, Kalika",
    booktitle = "Proceedings of the 2023 Conference on Empirical Methods in Natural Language Processing",
    month = dec,
    year = "2023",
    address = "Singapore",
    publisher = "Association for Computational Linguistics",
    url = "https://aclanthology.org/2023.emnlp-main.825/",
    doi = "10.18653/v1/2023.emnlp-main.825",
    pages = "13358--13376"
}

@inproceedings{jiang-etal-2024-longllmlingua,
    title = "{L}ong{LLML}ingua: Accelerating and Enhancing {LLM}s in Long Context Scenarios via Prompt Compression",
    author = "Jiang, Huiqiang  and
      Wu, Qianhui  and
      Luo, Xufang  and
      Li, Dongsheng  and
      Lin, Chin-Yew  and
      Yang, Yuqing  and
      Qiu, Lili",
    editor = "Ku, Lun-Wei  and
      Martins, Andre  and
      Srikumar, Vivek",
    booktitle = "Proceedings of the 62nd Annual Meeting of the Association for Computational Linguistics (Volume 1: Long Papers)",
    month = aug,
    year = "2024",
    address = "Bangkok, Thailand",
    publisher = "Association for Computational Linguistics",
    url = "https://aclanthology.org/2024.acl-long.91/",
    doi = "10.18653/v1/2024.acl-long.91",
    pages = "1658--1677"
}

@inproceedings{hu-etal-2025-hiagent,
    title = "{H}i{A}gent: Hierarchical Working Memory Management for Solving Long-Horizon Agent Tasks with Large Language Model",
    author = "Hu, Mengkang  and
      Chen, Tianxing  and
      Chen, Qiguang  and
      Mu, Yao  and
      Shao, Wenqi  and
      Luo, Ping",
    editor = "Che, Wanxiang  and
      Nabende, Joyce  and
      Shutova, Ekaterina  and
      Pilehvar, Mohammad Taher",
    booktitle = "Proceedings of the 63rd Annual Meeting of the Association for Computational Linguistics (Volume 1: Long Papers)",
    month = jul,
    year = "2025",
    address = "Vienna, Austria",
    publisher = "Association for Computational Linguistics",
    url = "https://aclanthology.org/2025.acl-long.1575/",
    doi = "10.18653/v1/2025.acl-long.1575",
    pages = "32779--32798",
    ISBN = "979-8-89176-251-0"
}

@inproceedings{zhou2026mem1,
  title={Mem1: Learning to synergize memory and reasoning for efficient long-horizon agents},
  author={Zhou, Zijian and Qu, Ao and Wu, Zhaoxuan and Kim, Sunghwan and Prakash, Alok and Rus, Daniela and Low, Bryan Kian Hsiang and Liang, Paul},
  booktitle={International Conference on Learning Representations},
  volume={2026},
  pages={58413--58438},
  year={2026}
}

@article{kang2025acon,
  title={Acon: Optimizing context compression for long-horizon llm agents},
  author={Kang, Minki and Chen, Wei-Ning and Han, Dongge and Inan, Huseyin A and Wutschitz, Lukas and Chen, Yanzhi and Sim, Robert and Rajmohan, Saravan},
  journal={arXiv preprint arXiv:2510.00615},
  year={2025}
}

@inproceedings{zhang2026memory,
  title={Memory as action: Autonomous context curation for long-horizon agentic tasks},
  author={Zhang, Yuxiang and Shu, Jiangming and Ma, Ye and Lin, Xueyuan and Wu, Shangxi and Sang, Jitao},
  booktitle={Findings of the Association for Computational Linguistics: ACL 2026},
  pages={19149--19164},
  year={2026}
}

@article{huang2024context,
  title={In-context decision transformer: Reinforcement learning via hierarchical chain-of-thought},
  author={Huang, Sili and Hu, Jifeng and Chen, Hechang and Sun, Lichao and Yang, Bo},
  journal={arXiv preprint arXiv:2405.20692},
  year={2024}
}

@article{zhou2026safedrive,
  title={Safedrive: Knowledge-and data-driven risk-sensitive decision-making for autonomous vehicles with large language models},
  author={Zhou, Zhiyuan and Huang, Heye and Li, Boqi and Zhao, Shiyue and Mu, Yao and Wang, Jianqiang},
  journal={Accident Analysis \& Prevention},
  volume={224},
  pages={108299},
  year={2026},
  publisher={Elsevier}
}

@inproceedings{fu2024drive,
  title={Drive like a human: Rethinking autonomous driving with large language models},
  author={Fu, Daocheng and Li, Xin and Wen, Licheng and Dou, Min and Cai, Pinlong and Shi, Botian and Qiao, Yu},
  booktitle={2024 IEEE/CVF Winter Conference on Applications of Computer Vision Workshops (WACVW)},
  pages={910--919},
  year={2024},
  organization={IEEE}
}

@inproceedings{zhang2024proagent,
  title={Proagent: building proactive cooperative agents with large language models},
  author={Zhang, Ceyao and Yang, Kaijie and Hu, Siyi and Wang, Zihao and Li, Guanghe and Sun, Yihang and Zhang, Cheng and Zhang, Zhaowei and Liu, Anji and Zhu, Song-Chun and others},
  booktitle={Proceedings of the AAAI Conference on Artificial Intelligence},
  volume={38},
  number={16},
  pages={17591--17599},
  year={2024}
}

@article{yan2025hybrid,
  title={Hybrid LLM-DDQN-based joint optimization of V2I communication and autonomous driving},
  author={Yan, Zijiang and Zhou, Hao and Tabassum, Hina and Liu, Xue},
  journal={IEEE Wireless Communications Letters},
  volume={14},
  number={4},
  pages={1214--1218},
  year={2025},
  publisher={IEEE}
}

@article{yuan2024rag,
  title={Rag-driver: Generalisable driving explanations with retrieval-augmented in-context learning in multi-modal large language model},
  author={Yuan, Jianhao and Sun, Shuyang and Omeiza, Daniel and Zhao, Bo and Newman, Paul and Kunze, Lars and Gadd, Matthew},
  journal={arXiv preprint arXiv:2402.10828},
  year={2024}
}

@inproceedings{wang2022scienceworld,
  title={Scienceworld: Is your agent smarter than a 5th grader?},
  author={Wang, Ruoyao and Jansen, Peter and C{\^o}t{\'e}, Marc-Alexandre and Ammanabrolu, Prithviraj},
  booktitle={Proceedings of the 2022 Conference on Empirical Methods in Natural Language Processing},
  pages={11279--11298},
  year={2022}
}

@inproceedings{wei-etal-2026-pace,
    title = "{PACE}: Predictive Adaptive Context Extraction for Long-Horizon {LLM} Agents",
    author = "Wei, Lei  and
      Peng, Xiao  and
      Tt  and
      Zhang, Guannan  and
      Jiang, Chenhao  and
      Li, Hongyu  and
      Lin, Lanbo  and
      Xu, Yuanwu  and
      Liu, Jiayao  and
      Wang, Kesu  and
      Wang, Bin",
    editor = "Liakata, Maria  and
      Moreira, Viviane P.  and
      Zhang, Jiajun  and
      Jurgens, David",
    booktitle = "Proceedings of the 64th Annual Meeting of the {A}ssociation for {C}omputational {L}inguistics (Volume 1: Long Papers)",
    month = jul,
    year = "2026",
    address = "San Diego, California, United States",
    publisher = "Association for Computational Linguistics",
    url = "https://aclanthology.org/2026.acl-long.1252/",
    doi = "10.18653/v1/2026.acl-long.1252",
    pages = "27184--27199",
    ISBN = "979-8-89176-390-6",
}

@article{deng2026influence,
  title={Influence guided context selection for effective retrieval-augmented generation},
  author={Deng, Jiale and Shen, Yanyan and Pei, Ziyuan and Chen, Youmin and Huang, Linpeng},
  journal={Advances in Neural Information Processing Systems},
  volume={38},
  pages={29225--29252},
  year={2026}
}

@article{ning2026agent,
  title={Agent-Omit: Adaptive Context Omission for Efficient LLM Agents},
  author={Ning, Yansong and Fang, Jun and Tan, Naiqiang and Liu, Hao},
  journal={arXiv preprint arXiv:2602.04284},
  year={2026}
}

\appendix

\section{Baselines and Evaluation Metrics}

\label{app:baseline}
We compare our method with three baselines: 
\begin{enumerate}
    \item Full H$_{20}$: Conditions on the entire current rolling window of up to 20 recent interactions at every decision, without selection or compression. It does not perform history selection or context compression and serves as the full-context reference.
    \item Similarity-K: keeps the $K$ history items that are most similar to the current state. Similarity is computed using the state representations of the current and historical observations, such as ego speed, lane information, nearby vehicles, conflict-related information, junction state, and traffic-light state for autonomous driving and actions taken, resulting observations, available objects, and rewards recorded in each historical attempt for ScienceWorld. Let $z_t$ and $z_i$ denote the feature vectors of the current state and historical state $h_i$, respectively. Let $e_t = E(q_t), e_i = E(h_i)$ be text embedding of query and history. Each feature is standardized as 
    \[
    \tilde{z}_{i,j} = \frac{z_{i,j}-\mu_j}{\sigma_j},
    \]
    where $\mu_j$ and $\sigma_j$ are the mean and standard deviation of feature $j$ from the dataset.

    We compute the similarity using Euclidean distance for autonomous driving:
    \[
    d_i =   \left\| \tilde{z}_i-\tilde{z}_t \right\|_2.
    \]
    and cosine distance for ScienceWorld:
    \[
    d_i =   1-\text{cos}(e_i,e_t).
    \]
    The $K$ histories with the smallest distances are selected:
    \[
    S_t^{\mathrm{sim}}  = \arg\min_{\substack{S\subseteq H_t,\; |S|=K}}    
    \sum_{h_i\in S} d_i.
    \]

    If ties, we will select the more recent chunk. This baseline tests whether similarity-based selection is sufficient for closed loop evaluation.
    
    \item Recent-K: keeps the $K$ most recent history items and removes the older histories. This baseline tests whether using a simple recency-based context is sufficient.
\end{enumerate}

Following prior work on recency-based memory retrieval ~\cite{zhang2024proagent} and state-similarity-based driving-experience retrieval, we implement Recent-$K$ and Similarity-$K$ as task-adapted baselines ~\cite{yan2025hybrid}. For Recent-$K$ and Similarity-$K$, we choose $K$ to approximately match the average number of history items kept by Ours based on results in each evaluation setting. This allows us to compare different context-selection strategies under a similar history budget.

We additionally compare our method with compression-controlled baselines: Recent-K + Redundancy Compression (Recent-K + LORC). These baselines retains the \(K\) most recent history items and applies the same learned order-guided redundancy compression used by Ours. 

Furthermore, we compare our method to a related context management benchmark, LongLLMLingua, on ScienceWorld \cite{jiang-etal-2024-longllmlingua}. We calibrate its compression target to the average post-H20 historical-context budget of Ours and fix the resulting compression setting before the formal evaluation. Because LongLLMLingua and the downstream policy use different tokenizers, we report average total context usage using the downstream tokenizer across all trials. 

\section{Order-Guided Redundancy Compression}
\label{app:lorc}
We divide the information in each history item into two groups. Decision-specific dynamic information, such as the previous decision, action, and reward, is always retained. For ScienceWorld, these fields include the observation text and available-object list within each retained attempt summary. For autonomous driving, we apply the operation only to predefined contextual fields that can repeat across retained interaction histories.

Let $v_i^f$ denote the value of compressible field $f$ in selected
history item $h_i$, and let $v_t^f$ denote the corresponding value in
the current query. We process each contextual value according to where
the same information is already represented. If $v_i^f=v_t^f$, the
historical occurrence can be deleted because the complete value is already available there. If a value $v$ is not present in the current query and occurs in only one retained history, it is left unchanged. If the same value $v$ appears in multiple selected histories, we keep one complete occurrence.

The retained history that keeps the complete occurrence is selected
using the deletion order:
\begin{equation}
    c_{t,f,v}
=
\arg\max_{i:\,h_i\in S_{K_t^*},\,v_i^f=v}
\rho_{t,i},
\label{eq:reference_location}
\end{equation}
where $\rho_{t,i}$ is the position of $h_i$ in the deletion order.
Because a larger $\rho_{t,i}$ means that $h_i$ is removed later,
$h_{c_{t,f,v}}$ is the retained history with the largest deletion-order position among those containing value $v$. We therefore keep the
complete value in $h_{c_{t,f,v}}$ and replace its other occurrences
with references to this retained history based on environment-specific compression rule. In this way, the same deletion order used to select histories also determines which retained history stores each repeated contextual value.

The final policy prompt is therefore
\[
P_t
=
\Gamma
\left(
q_t,S_{K_t^*},\rho_t
\right).
\]
where $K_t^*$ is the adaptively selected history size and $\Gamma$ constructs the final prompt using the deletion order.

Two deletion orderings can produce the same retained subset but assign repeated values to different chunks, resulting in different final prompts. 

\section{Task-Specific Representation}
\label{app:representation}
In autonomous driving, the candidate representation incorporates the current deletion candidate and driving decision information; For a candidate deletion, let $P$ denote the current parent context and $S$ denote the resulting selected context. We compute their average representations as
\[
\bar{u}_P
=
\frac{1}{|P|}
\sum_{i \in P} u_i,
\qquad
\bar{u}_S
= 
\frac{1}{|S|}
\sum_{i \in S} u_i.
\]
In ScienceWorld, it summarizes the selected, excluded, and full attempt histories. The predictor serves the same role in both settings: estimating the effect of retaining $S$ instead of the full history.

\section{Predictor Training}
\label{app:predictor}
Our predictor uses a five-member predictor ensemble. Every member contains a
one-layer Transformer encoder with four attention heads, hidden dimension 64,
and dropout 0.1. Predictors are optimized with AdamW at a learning rate of
$3\times10^{-4}$ and gradient clipping at 1.0. The training objective combines
absolute Huber regression, same-parent (or same-cardinality) difference
regression, and pairwise ranking with weights $1$, $0.5$, and $0.25$,
respectively. Driving uses weight decay $10^{-4}$ and ScienceWorld uses weight
decay $10^{-2}$. Checkpoint selection is performed only on the reserved
development split and is separate from predictor weight fitting.

\begin{table*}[!t]
\centering
\caption{
Comparison of Recent, Recent+LORC, and Ours across evaluation scenarios.
For driving, total tokens denote post-selection tokens and performance is
Safety/Comfort/Efficiency (S/C/E). For ScienceWorld, total tokens denote
post-selection tokens per trial and performance is return.
Lower token usage is better; higher performance is better.
}
\label{tab:scenario_results}

\begingroup
\scriptsize
\renewcommand{\arraystretch}{0.95}

\begin{tabular*}{\textwidth}{
@{\extracolsep{\fill}}
llcccc
@{}}
\toprule
\textbf{Scenario}
& \textbf{Method}
& \textbf{Mean $K$}
& \textbf{Post-selection Tokens$\downarrow$}
& \textbf{Performance$\uparrow$} \\
\midrule

\multirow{3}{*}{Held-out ID}
& Recent-13
& $13.00$
& $91{,}730$
& $\mathbf{0.8838}/0.7617/0.7981$ \\

& Recent-13 + LORC
& $13.00$
& $74{,}458$
& $0.8808/0.7602/0.7825$ \\

& \textbf{Ours}
& $13.63$
& $\mathbf{73{,}233}$
& $0.8836/\mathbf{0.7645}/\mathbf{0.8071}$ \\

\midrule

\multirow{3}{*}{Severe-3x}
& Recent-13
& $13.00$
& $102{,}971$
& $\mathbf{0.8673}/0.7715/\mathbf{0.8351}$ \\

& Recent-13 + LORC
& $13.00$
& $98{,}165$
& $0.8561/\mathbf{0.7802}/0.7996$ \\

& \textbf{Ours}
& $13.02$
& $\mathbf{85{,}747}$
& $0.8625/0.7759/0.8256$ \\

\midrule

\multirow{3}{*}{Unseen Routes}
& Recent-15
& $15.00$
& $95{,}859$
& $\mathbf{0.9828}/\mathbf{0.7457}/\mathbf{0.8803}$ \\

& Recent-15 + LORC
& $15.00$
& $81{,}468$
& $0.9771/0.7401/0.8598$ \\

& \textbf{Ours}
& $15.05$
& $\mathbf{75{,}138}$
& $0.9827/0.7412/0.8706$ \\

\midrule

\multirow{3}{*}{ScienceWorld}
& Recent-12
& $12.00$
& $202{,}710$
& $\mathbf{96.06}$ \\

& Recent-12 + LORC
& $12.00$
& $94{,}254$
& $94.78$ \\

& \textbf{Ours}
& $12.70$
& $\mathbf{87{,}100}$
& $\mathbf{96.06}$ \\

\bottomrule
\end{tabular*}

\endgroup
\end{table*}

\section{Controlled Baseline Analysis}
\label{app:redundancy}
\textbf{Redundancy Compression Controlled Recency Baseline}
To isolate the effect of learned subset selection, we additionally apply the same learned order-guided redundancy compression (LORC) to the Recent-\(K\) baseline. We report post-selection total tokens to isolate the selection-active phase as shown in table \ref{tab:scenario_results}.

The benefit is most pronounced under distribution shift. Relative to Recent-$K$+LORC, Ours reduces post-selection total tokens by $12.7\%$ on Severe-3x and $7.8\%$ on unseen routes, respectively. On held-out ID, the controlled gap is smaller. The larger gap under the two OOD settings suggests that downstream-aware history selection is particularly useful when simple recency is less aligned with the context needed for the current decision. 

A similar controlled advantage appears in ScienceWorld. Ours reduces post-selection total tokens by $7.6\%$ and average total tokens by $7.4\%$ relative to Recent-12+LORC. Notably, applying LORC to Recent-12 lowers its observed running-max return from $96.06$ to $94.78$, whereas Ours retains a return of $96.06$ while using fewer tokens. This indicates that redundancy compression is not performance-neutral when applied to a fixed recency-selected subset, and that its interaction with history selection matters. Together, these controlled comparisons show that the gains over recency-based selection are not explained by redundancy compression alone.

\textbf{Matched fixed-budget Baseline}
To isolate the effect of online budget determination, we additionally compare our method with fixed K baseline which follows the exactly same method as ours. For each setting, the fixed $K$ is chosen to match the mean retained history size observed for Ours. At matched context budget, the fixed and online-budget variants achieve similar operating points on held-out ID and unseen routes. On Severe-3x, Ours
reduces post-selection total tokens by 7.4\% relative to Fixed-$K$13, with comparable driving performance. These results are intended as a controlled comparison of budget determination. 

\textbf{Budget Determination}
The fixed-budget variants require a separately specified \(K\) for each setting. We calibrate these values to the mean context sizes produced by our adaptive method. Our method achieves comparable performance using a selection threshold, without separately specifying a fixed \(K\) for each setting. The comparison therefore highlights the ability to determine context budgets automatically, alongside the setting-dependent additional savings from decision-dependent budgeting.

\textbf{Decision-Level Budget Variation} Although the adaptive and matched fixed-$K$ variants have similar mean context budget, their decision-level budgets differ substantially. Across ID, Severe-3x, and unseen routes, 89.4\%, 89.7\%, and 88.1\% of selection-active decisions, respectively, differ from the corresponding matched fixed $K$ by at least two history items. The variation also occurs within individual trajectories: on unseen routes, all 27 episodes show variation in $K$, and the retained budget changes between 64.9\% of selection-active decisions.

\begin{table*}[!t]
\centering
\caption{
Sensitivity to the feasibility threshold $\tau$.
Post-selection input tokens are estimated from the 21st decision onward
on fixed trajectories and averaged per episode.
Changes are reported relative to $\tau=0.05$.
}
\label{tab:tau_sensitivity}
\vspace{-4pt}

\begingroup
\scriptsize
\setlength{\tabcolsep}{5pt}
\renewcommand{\arraystretch}{0.78}
\setlength{\aboverulesep}{1pt}
\setlength{\belowrulesep}{1pt}

\begin{tabular*}{0.92\textwidth}{
@{\extracolsep{\fill}}
lcccc
@{}}
\toprule
\textbf{Setting}
& $\boldsymbol{\tau}$
& \textbf{Mean $K$}
& \textbf{Estimated Post-selection Input Tokens$\downarrow$}
& \textbf{Change vs. $\tau=0.05$} \\
\midrule

\multirow{6}{*}{Held-out ID}
& $0.02$
& $17.58$
& $81{,}734$
& $+17.79\%$ \\

& $0.03$
& $16.50$
& $78{,}402$
& $+12.99\%$ \\

& $\mathbf{0.05}$
& $\mathbf{13.63}$
& $\mathbf{69{,}387}$
& $\mathbf{0.00\%}$ \\

& $0.07$
& $11.64$
& $63{,}183$
& $-8.94\%$ \\

& $0.10$
& $10.64$
& $59{,}685$
& $-13.98\%$ \\

& $0.15$
& $10.14$
& $57{,}931$
& $-16.51\%$ \\

\midrule

\multirow{6}{*}{Severe-3x}
& $0.02$
& $17.63$
& $99{,}111$
& $+22.28\%$ \\

& $0.03$
& $16.62$
& $94{,}618$
& $+16.74\%$ \\

& $\mathbf{0.05}$
& $\mathbf{13.02}$
& $\mathbf{81{,}050}$
& $\mathbf{0.00\%}$ \\

& $0.07$
& $11.26$
& $74{,}569$
& $-8.00\%$ \\

& $0.10$
& $10.47$
& $71{,}851$
& $-11.35\%$ \\

& $0.15$
& $10.15$
& $70{,}677$
& $-12.80\%$ \\

\midrule

\multirow{6}{*}{Unseen Routes}
& $0.02$
& $19.62$
& $83{,}729$
& $+18.26\%$ \\

& $0.03$
& $18.85$
& $81{,}569$
& $+15.21\%$ \\

& $\mathbf{0.05}$
& $\mathbf{15.05}$
& $\mathbf{70{,}803}$
& $\mathbf{0.00\%}$ \\

& $0.07$
& $12.03$
& $62{,}174$
& $-12.19\%$ \\

& $0.10$
& $10.39$
& $57{,}455$
& $-18.85\%$ \\

& $0.15$
& $10.00$
& $56{,}429$
& $-20.30\%$ \\

\bottomrule
\end{tabular*}

\endgroup
\vspace{-6pt}
\end{table*}
\begin{wraptable}{r}{0.58\textwidth}
\vspace{-10pt}
\centering
\caption{ScienceWorld Post-selection return comparison. Higher is better.}
\label{tab:sw_post_selection_return}
\vspace{-5pt}

\begingroup
\scriptsize
\renewcommand{\arraystretch}{0.92}
\setlength{\tabcolsep}{4pt}

\begin{tabular*}{\linewidth}{@{\extracolsep{\fill}}lcc@{}}
\toprule
\textbf{Method}
& \textbf{Mean Return$\uparrow$}
& \textbf{Running-Max Return$\uparrow$} \\
\midrule

Full-H20
& $74.34$
& $95.50$ \\

Recent-12
& $\mathbf{76.02}$
& $95.78$ \\

Similarity-12
& $69.37$
& $94.78$ \\

Recent-12 + LORC
& $70.27$
& $94.78$ \\

LongLLMLingua (matched-budget)
& $73.74$
& $94.67$ \\

\textbf{Ours}
& $74.22$
& $\mathbf{95.94}$ \\

\bottomrule
\end{tabular*}

\endgroup
\vspace{-8pt}
\end{wraptable}

\section{ScienceWorld Performance across Rounds}
\label{app:sw_performance}
\begin{figure}[t]
    \centering
    \includegraphics[width=0.72\textwidth]
    {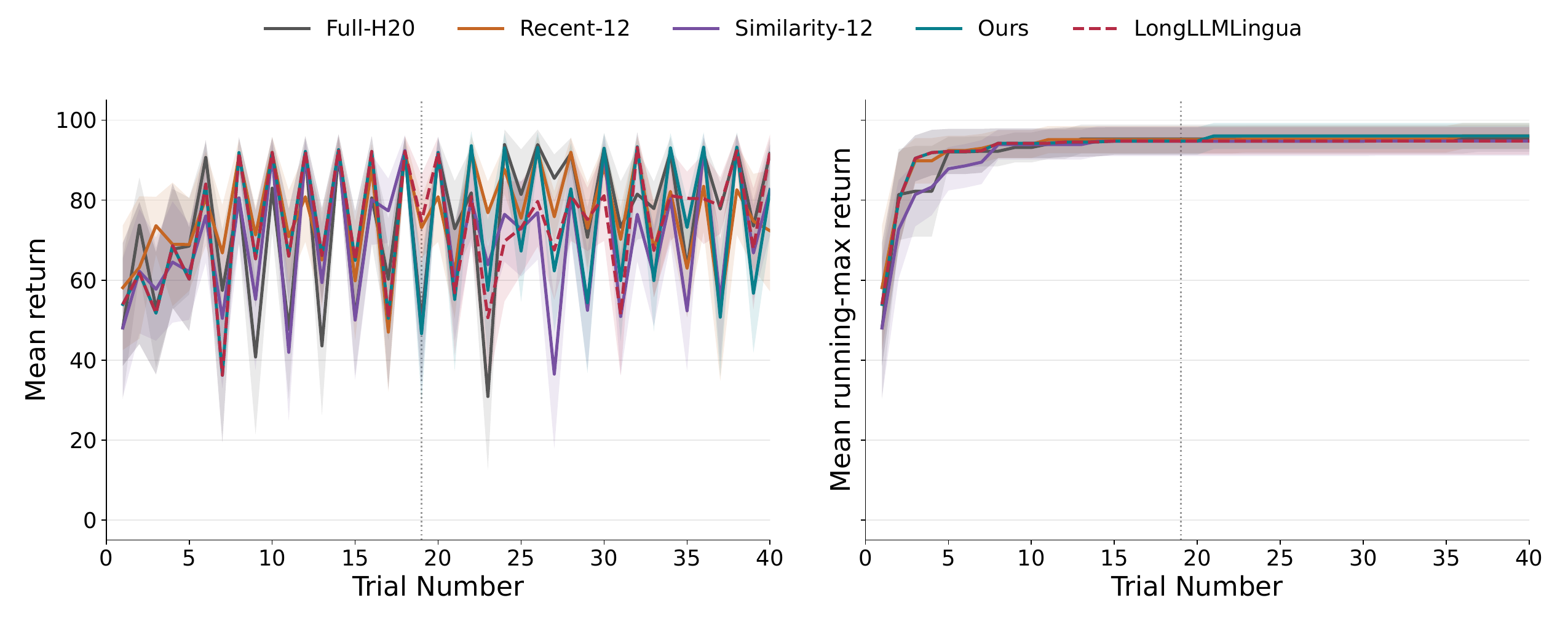}
    \caption{ScienceWorld performance over 40 interaction trials.
    Left: mean trial return. Right: mean running-max return.
    Shaded regions denote SEM across tasks.}
    \label{fig:sw_return}
\end{figure}

Fig.\ref{fig:sw_return} reports the mean return at each formal trial and the corresponding running maximum across the 40 interaction rounds. While trial-level returns exhibit variation across tasks and rounds, the running-max curves converge to similar high-return levels for all methods. This analysis complements the running-max return reported in Table \ref{tab:main_results} by showing the evolution of observed performance throughout the interaction process. Additionally, we report the running-max/mean return over post-selection period (Round 19-40) in Table \ref{tab:sw_post_selection_return}. Ours achieves 95.94/74.22, compared with 95.78/76.02 for Recent, 94.78/69.37 for Similarity, 94.78/70.27 for Recent+LORC, and 94.67/73.74 for LongLLMLingua.

\section{LongLLMLingua Baseline and Matched-Budget}
\label{app:longllm}
\textbf{Implementation and budget matching.}
We use the official LongLLMLingua v0.2.2 implementation without modifying its
compression core \cite{jiang-etal-2024-longllmlingua}. The compressor is calibrated to the average post-H20
historical-context budget of Ours, measured using the downstream
\texttt{o200k\_base} tokenizer. Because LongLLMLingua uses its own tokenizer
internally, this calibration yields a fixed compressor-side target of 9,042
tokens. The target is fixed before formal evaluation, and all reported token
usage is recomputed using the downstream tokenizer. LongLLMLingua is evaluated
under the same ScienceWorld protocol as the other methods.

\textbf{Latency measurement.}
For Ours, context-management latency is measured from history embedding
generation through adaptive selection, redundancy compression, final prompt, and
validation. For LongLLMLingua, we time the official compression core after
model warm-up, with GPU synchronization immediately before and after each
compression call. Both measurements exclude model loading, network
communication, downstream policy inference, and environment execution.
LongLLMLingua processes raw histories of approximately 31.8k
compressor-tokenizer tokens on average. The reported timings characterize the
observed deployment overhead and are not hardware-normalized speed comparisons.

\section{Feasibility Threshold Sensitivity Analysis}
\label{app:sensitivity}
We analyze the sensitivity of the adaptive context budget to the feasibility threshold \(\tau\). Using the recorded predictor outputs from all 1,114 post-H20 driving decisions, we recompute context selection offline for multiple threshold values while keeping the original evaluation trajectories fixed. For each \(\tau\), we report the resulting mean selected context size \(K\) and the post-selection input tokens accumulated from the 21st decision onward and averaged across episodes as shown in Table \ref{tab:tau_sensitivity} 

\section{Paired-bootstrap confidence intervals for performance}
\label{app:ci}
We additionally report paired-bootstrap 95\% confidence intervals for the
performance differences between Ours and Full-H20, with
$\Delta=\text{Ours}-\text{Full-H20}$. Bootstrap resampling is performed
over matched route--domain--seed episodes. On held-out ID, the
differences are $+0.0060$ for Safety (95\% CI $[-0.0090,+0.0208]$),
$+0.0065$ for Comfort ($[-0.0027,+0.0160]$), and $+0.0132$ for Efficiency
([$+0.0020,+0.0252$]). On Severe-3x, the corresponding differences
are $+0.0414$ ($[-0.0013,+0.0963]$), $+0.0000$
($[-0.0168,+0.0156]$), and $-0.0012$ ($[-0.0226,+0.0192]$).
On unseen routes, they are $-0.0007$ ($[-0.0039,+0.0024]$), \\
$-0.0005$ ($[-0.0132,+0.0124]$), and $+0.0018$
($[-0.0111,+0.0137]$) for Safety, Comfort, and Efficiency, respectively. For ScienceWorld, we perform paired bootstrap resampling over the tasks. The mean-return 95\% confidence interval is $[-5.0,+4.3]$. All driving confidence intervals include zero except for the positive Efficiency difference on held-out ID, while the ScienceWorld mean-return interval also includes zero.
Overall, we find no statistically detectable degradation in any driving or ScienceWorld
metric relative to Full-H20.
\end{document}